\pdfoutput=1
\documentclass[11pt]{article}

\usepackage{acl}

\usepackage{times}
\usepackage{latexsym}

\usepackage[T1]{fontenc}
\usepackage[utf8]{inputenc}

\usepackage{microtype}

\usepackage{amsmath}
\usepackage{bbm}
\usepackage{bm}
\usepackage{booktabs}
\usepackage{subfigure}
\usepackage{multirow}
\usepackage{color}
\usepackage{graphicx}
\usepackage{makecell}
\usepackage{amssymb}
\usepackage{tikz}
\usepackage{pgfplots}

\title{Bridging the Domain Gaps in Context Representations for $k$-Nearest Neighbor Neural Machine Translation}

\author{Zhiwei Cao\textsuperscript{1,3}\thanks{\hspace{2mm}This work was done when Zhiwei Cao was interning at DAMO Academy, Alibaba Group.},~~\textbf{Baosong Yang}\textsuperscript{2},~~\textbf{Huan Lin}\textsuperscript{2},~~\textbf{Suhang Wu}\textsuperscript{1},~~\textbf{Xiangpeng Wei}\textsuperscript{2} \\
\textbf{Dayiheng Liu}\textsuperscript{2},~~\textbf{Jun Xie}\textsuperscript{2},~~\textbf{Min Zhang}\textsuperscript{4} and \textbf{Jinsong Su}\textsuperscript{1,3}\thanks{\hspace{2mm}Corresponding author.} \\
\textsuperscript{1}School of Informatics, Xiamen University, China \\
\textsuperscript{2}Language Technology Lab, Alibaba DAMO Academy \\
\textsuperscript{3}Institute of Artificial Intelligence, Xiamen University, China \\
\textsuperscript{4}Institute of Computer Science and Technology, Soochow University, China \\
\texttt{lines1@stu.xmu.edu.cn, jssu@xmu.edu.cn}
}
\begin{document}
\maketitle

\begin{abstract}
$k$-Nearest neighbor machine translation ($k$NN-MT) has attracted increasing attention due to its ability to non-parametrically adapt to new translation domains. By using an upstream NMT model to traverse the downstream training corpus, it is equipped with a datastore containing vectorized key-value pairs, which are retrieved during inference to benefit translation.
However, there often exists a significant gap between upstream and downstream domains, which hurts the retrieval accuracy and the final translation quality.
To deal with this issue, we propose a novel approach to boost the datastore retrieval of $k$NN-MT by reconstructing the original datastore.
Concretely, we design a reviser to revise the key representations, making them better fit for the downstream domain. The reviser is trained using the collected semantically-related key-queries pairs, and optimized by two proposed losses: one is the key-queries semantic distance ensuring each revised key representation is semantically related to its corresponding queries, and the other is an L2-norm loss encouraging revised key representations to effectively retain the knowledge learned by the upstream NMT model. 
Extensive experiments on domain adaptation tasks demonstrate that our method can effectively boost the datastore retrieval and translation quality of $k$NN-MT.\footnote{Our code is available at \url{https://github.com/DeepLearnXMU/RevisedKey-knn-mt}.}
\end{abstract}

\section{Introduction}
\label{sec:introduction}
\begin{figure}[t!]
\centering
\includegraphics[width=0.98\linewidth]{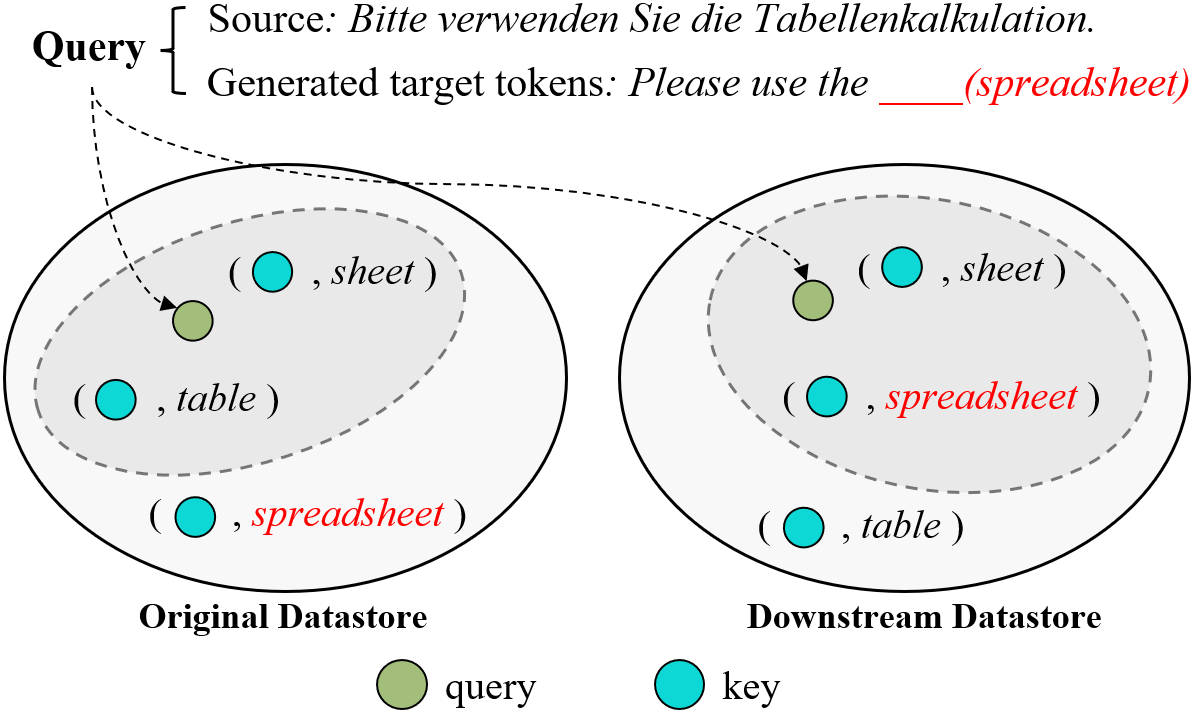}
\caption{An example of datastore retrieval, where News and IT are the upstream and downstream domains, respectively. We first build a downstream NMT model by fine-tuning the upstream NMT model on the downstream training corpus. Then, we use the downstream NMT model to re-traverse the downstream training corpus, constructing a downstream datastore. Finally, we reuse the upstream and downstream NMT model to conduct retrieval on the original and downstream datastores, respectively. The result shows that the nearest neighbors retrieved by the same query are quite different, and only the retrieved nearest neighbors from the downstream datastore contain the ground-truth token ``\emph{spreadsheet}''.}
\label{fig:example}
\vspace{-4mm}
\end{figure}
The recently proposed $k$-Nearest Neighbors Machine Translation ($k$NN-MT)~\cite{khandelwal2021nearest} is increasingly receiving attention from the community of machine translation due to its advantage on non-parametric domain adaptation~\cite{zheng-etal-2021-adaptive, wang-etal-2022-efficient, meng-etal-2022-fast}. Given an \emph{upstream NMT model}, $k$NN-MT first uses the downstream training corpus to establish a datastore containing key-value pairs, where each key is the representation of the NMT decoder and its value is the corresponding target token. During inference, it uses the current decoder representation as a query to retrieve $N_k$ nearest key-value pairs from the datastore.
Afterwards, the retrieved values are transformed into a probability distribution based on the query-key distances, denoted as $k$NN distribution. Finally, this distribution is interpolated with the prediction distribution of the NMT model to adjust the prediction translation. By doing so, the upstream NMT model can be easily adapted to diverse domains by equipping domain-specific datastores without additional parameters. To avoid confusion in subsequent descriptions, we name the datastore in conventional $k$NN-MT as the \emph{original datastore}. 

However, there often exists a significant domain gap between the upstream NMT model and the downstream training corpus~\cite{koehn-knowles-2017-six, hu-etal-2019-domain-adaptation}. The learned key representations of the original datastore deviate from the ideal distribution of downstream-domain key representation. 
As shown in Figure~\ref{fig:example}, in the original datastore built by the News domain NMT model, the nearest neighbors of the query contain the out-domain token ``\emph{table}'' rather than the target token ``\emph{spreadsheet}'' from the IT domain.
This hurts the datastore retrieval of $k$NN-MT.
To alleviate the negative impact of the retrieval error, previous studies resort to dynamically estimating the weight of $k$NN distribution for the final prediction~\cite{zheng-etal-2021-adaptive, jiang-etal-2021-learning, jiang2022towards}. However, these studies ignore the key representation learning, which is the basis of constructing datastore, and low-quality key representations tend to result in retrieval errors.

To bridge the domain gap, a natural choice is to fine-tune the NMT model on the downstream training corpus to obtain the \emph{downstream NMT model} and then use it to build a $\emph{downstream datastore}$. 
However, this method has two serious defects: 1) it is required to deploy multiple domain-specific NMT models when dealing with multi-domain translations, involving huge system deployment overhead. For example, in the commonly-used $k$NN-MT datasets~\cite{aharoni-goldberg-2020-unsupervised} involving four downstream domains, this method has to construct four NMT models with datastores, consuming 37.2G GPU memory with 1,028M parameters. By contrast, $k$NN-MT involves only one NMT model and four datastores, consuming 11.3G GPU memory with 257M parameters;
2) it tends to be affected by the notorious catastrophic forgetting problem, weakening the adaptability of $k$NN-MT.
This may result from the fine-tuned NMT model tending to forget previous upstream-domain knowledge and are therefore challenging to adapt to other domains.
Thus, how to make more effective domain adaptation using $k$NN-MT remains a problem worth exploring.

In this paper, we propose a novel approach to boost the datastore retrieval of $k$NN-MT by reconstructing the original datastore.
Concretely, we design a \emph{Key Representation Reviser} that revises the key representations in an offline manner, so that they can better adapt to the retrieval from the downstream domain. This reviser is a two-layer feed-forward (FFN) with a ReLU function, which is fed with the information about a key representation $k$, and outputs an inductive bias $\Delta k$ to revise $k$ as $\hat{k} = k + \Delta k$. To train the reviser, 
we first use the downstream NMT model to extract semantically-related key-queries pairs from the downstream datastore, and then use their counterparts in the upstream NMT model and original datastore as supervision signals of the reviser. For each key-queries pair, we introduce two training losses to jointly optimize the reviser:
1) the \emph{semantic distance loss}, which encourages each revised key representation to be adjacent to its semantically-related queries; 2) the \emph{semantic consistency loss}, which avoids the revised key representation to be far from the original one, and thus preserving the knowledge learned by the upstream NMT model.

To summarize, our contributions are as follows:
\begin{itemize}
\vspace{-2.2mm}
\item Through in-depth analysis, we reveal that the issue of the domain gap in $k$NN-MT hurts the effectiveness of the datastore retrieval.
\vspace{-2.2mm}
\item We propose a novel method to boost the datastore retrieval of $k$NN-MT by revising the key representations. To the best of our knowledge, our work is the first attempt to revise key representations of the $k$NN-MT datastore in an offline manner.
\vspace{-2.2mm}
\item Extensive experiments on a series of translation domains show that our method can strengthen the domain adaptation of $k$NN-MT without additional parameters during inference.
\end{itemize}

\section{Preliminary Study}
In this section, we first briefly introduce \emph{k}NN-MT~\cite{khandelwal2021nearest}, and then conduct a group of experiments to study the domain gap in $k$NN-MT.

\subsection{$k$NN-MT}
The construction of a $k$NN-MT model involves two key steps: using the downstream training corpus to create a datastore, and conducting translation with the help of the datastore.

\paragraph{Datastore Creation}
The common practice is to first use the upstream NMT model to traverse a downstream training corpus, where the decoder autoregressively extracts the contextual representations and corresponding target tokens to build a datastore. Specifically, for each bilingual sentence $(x, y)$ from the downstream training corpus $\mathcal{C}_{prime}$, the NMT model generates the contextual representation $f(x, y_{<t})$ of the $t$-th target token $y_t$ condition on both source sentence $x$ and preceding target tokens $y_{<t}$. Then, the key-value pair $(f(x, y_{<t}), y_t)$ will be added to the original datastore $(\mathcal{K}, \mathcal{V})$.

\paragraph{Translation with $k$NN Distribution} 
During translation, the decoder outputs a probability distribution $p_{\textrm{NMT}}(\hat{y}_{t}|x, \hat{y}_{<t})$ at each timestep $t$, where $\hat{y}_{<t}$ represents the previously-generated target tokens. Then, the decoder outputs the contextual representation $f(x, \hat{y}_{<t})$ as the query to retrieve the datastore $(\mathcal{K}, \mathcal{V})$, obtaining $N_k$ nearest key-value pairs according to the query-key $l_2$ distance. Denote the retrieved pairs as $\mathcal{R}$, the $k$NN distribution is computed as follows:
\begin{align}
p_{\textrm{kNN}}&(\hat{y}_t |x, \hat{y}_{<t}) \propto \\
&\sum_{\mathcal{R}}\mathbbm{1}_{\hat{y}_t = v_i} \exp (\frac{-d(k_i, f(x, \hat{y}_{<t}))}{T}), \nonumber
\label{eq:knn_prob}
\end{align}
\noindent where $T$ is the softmax temperature and $d(\cdot, \cdot)$ is the $L_2$ distance function. Finally, the predictive probability of $\hat{y}_t$ is defined as the interpolation of the decoder predictive probability and the \emph{k}NN distribution probability:
\begin{align}
p(\hat{y}_t|x, \hat{y}_{<t}) &= \lambda \cdot p_{\textrm{kNN}}(\hat{y}_t|x, \hat{y}_{<t}) \\ 
&+ (1-\lambda) \cdot p_{\textrm{NMT}} (\hat{y}_t|x, \hat{y}_{<t}),
\nonumber
\label{equ:complete_prob}
\end{align}

\noindent where $\lambda \in [0,1]$ is a fixed interpolation weight.

\subsection{The Domain Gap in $k$NN-MT}
\label{sec:domain-shift}
As mentioned previously, the performance of $k$NN-MT depends heavily on the quality of its datastore, which directly affects the datastore retrieval of the NMT model. However, the datastore key representations are provided by the upstream NMT model without considering the downstream information. Therefore, it is difficult for the upstream NMT model to effectively retrieve the key-value pairs related to the downstream domain, and thus negatively affect the subsequent translation prediction. 

To verify this conjecture, we conduct a group of experiments on the development sets of four downstream domains, of which details are provided in Section~\ref{sec:dataset-setting}. Concretely, we first construct two $k$NN-MT models: 1) \textbf{$k$NN-MT}. It is a vanilla $k$NN-MT model, which uses the upstream NMT model to traverse the downstream training corpus, forming an original datastore; 2) \textbf{$k$NN-MT(F)}. We first fine-tune the upstream NMT model on the downstream training corpus to obtain a downstream NMT model, and then use it to build a downstream datastore on the training corpus above. Apparently, compared with the conventional $k$NN-MT model, $k$NN-MT(F) is less affected by the domain gap and its key representations are more in line with the ideal distribution of downstream-domain key representation. Afterwards, we adopt the above two models to traverse the development sets of four downstream domains, where the decoder contextual representations are used to retrieve the corresponding datastores\footnote{During this process, we skip some meaningless tokens, like stopwords.}, respectively.

\begin{figure}[t]
\centering
\includegraphics[width=0.95\linewidth]{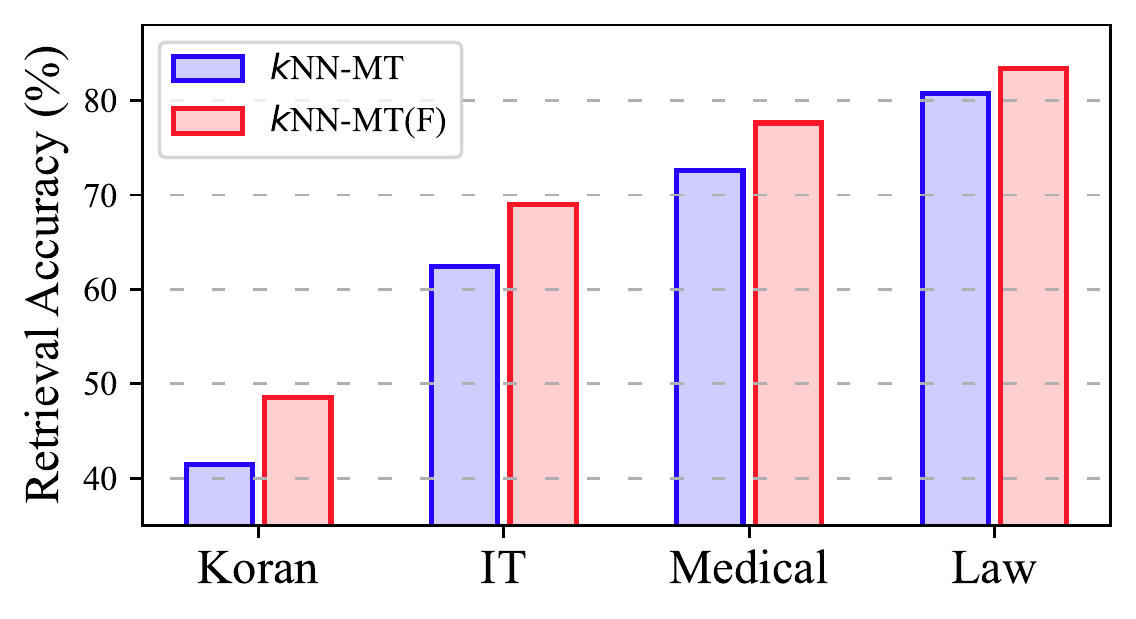}
\vspace{-3mm}
\caption{The retrieval accuracy of the conventional $k$NN-MT model on the original datastore, and $k$NN-MT(F) on the downstream datastore.}
\label{fig:retrieval-accuracy}
\end{figure}
To measure the retrieval quality of an NMT model on a datastore, we focus on those words retrieved with the maximal probability and define the proportion of ground-truth words in them as \emph{retrieval accuracy}.
Figure~\ref{fig:retrieval-accuracy} illustrates the retrieval accuracy of the above $k$NN-MT models. We have two important findings. First, $k$NN-MT(F) achieves higher retrieval accuracy than the conventional $k$NN-MT model in all domains. These results demonstrate that alleviating the domain gap can improve the datastore retrieval of $k$NN-MT; Second, although $k$NN-MT(F) is more suitable for the downstream domain, it is not perfect and there are still some retrieval errors.

Although $k$NN-MT(F) can achieve higher retrieval accuracy, it still suffers from huge system deployment overhead for multi-domain translation and catastrophic forgetting, as mentioned previously. To avoid these issues, we explore a trade-off solution that directly revises the key representations of the original datastore, so as to enhance the retrieval effectiveness for the conventional $k$NN-MT model.

\section{Our Method}
To alleviate the influence of the domain gap on the datastore retrieval of $k$NN-MT, we propose a simple yet effective approach to directly revise the original datastore, of which revised key representations are required to satisfy two properties: 1) they are more in line with the ideal distribution of downstream-domain key representation; 2) they can effectively retain the translation knowledge learned by the upstream NMT model.

To this end, we design a \emph{Key Representation Reviser} to revise the key representations of the original datastore. 
To train this reviser, we first identify some key-queries pairs from the original datastore and upstream NMT model as the training data, where each key is expected to be semantically related to its corresponding queries.
Then, we propose two training losses to jointly train the reviser.
Using the reviser to reconstruct the original datastore, the original datastore can also effectively capture the semantically related key-queries pairs contained in the downstream datastore and NMT model, and thus is more suitable for the downstream translation task.

\subsection{Key Representation Reviser}
\label{sec:reviser}
Our reviser is a two-layer FFN with a ReLU function. It is not embedded into the $k$NN-MT model, but can be used to modify key representations in an offline manner. For each key-value pair $(k, v)$ in the original datastore, we obtain its corresponding counterpart $(k', v)$ from the downstream datastore\footnote{Given the same source sentence $x$ and preceding target tokens $y_{<t}$, the key representation $k$ and $k'$ generated by upstream and downstream NMT models correspond to each other.}, and feed them into the reviser to generate an \emph{inductive bias vector} $\Delta k$ for revising $k$:
\begin{align}
& \Delta k = \text{FFN}([k; k'; \text{Emb}(v); \text{Emb}'(v)]), \\
& \hat{k} = k + \Delta k,
\end{align}
\noindent where $\hat{k}$ denotes the revised key representation, $\text{Emb}(\cdot)$ and $\text{Emb}'(\cdot)$ are the token embeddings of the upstream and the downstream NMT models, respectively.

\subsection{Training Data Construction}
\label{sec:construction}
To train the key representation reviser, we adopt three steps to construct training data. Specifically, we first use the downstream NMT model to extract semantically-related key-queries pairs from the downstream datastore. Then, we filter some extracted low-quality key-queries pairs. Finally, from the original datastore and the upstream NMT model, we determine the corresponding counterparts of the above-mentioned key-queries pairs as the training data. Next, we introduce these three steps in detail.

\textbf{Step 1}. As implemented in the previous preliminary study, we first construct a downstream NMT model $\theta '$ and its corresponding downstream datastore $\mathcal{D}'$. 
Then, we use the model $\theta '$ to re-traverse the downstream training corpus $\mathcal{C}_{prime}$, where the decoder representation is used as the query $q'$ to retrieve $N_k$ nearest key-value pairs $\{(k', v)\}$ from $\mathcal{D}'$.
In this process, we collect these queries and their corresponding key-value pairs from the datastore.
By doing so, we can easily determine a subset $\{q'\}$ corresponding to each $k'$ from all queries, and further obtain a set of semantically-related key-queries pairs.

\textbf{Step 2}. As mentioned in Section~\ref{sec:domain-shift}, the downstream datastore is not perfect. Thus, the above key-queries pairs may contain noise.

To alleviate this issue, we learn from the related studies \cite{tomasev2013role, he-etal-2021-efficient}, and filter the low-quality key-queries pairs according to the retrieval numbers of keys. As analyzed in \cite{he-etal-2021-efficient}, in high-dimensional data, a data point is considered more reliable if it belongs to the nearest neighbors of many other data points. Inspired by this, we count the retrieved numbers $\text{Count}(k')$ of each key $k'$ to measure its reliability. However, the keys with high-frequency values are originally retrieved more frequently. Only considering $\text{Count}(k')$ may result in some unreliable keys with high-frequent values being retained while some reliable pairs with low-frequent values being excluded. Therefore, we normalize $\text{Count}(k')$ with the token frequency $\text{Freq}(v)$ of its corresponding value $v$, and finally select the top $r$\% key-queries pairs sorted by $\text{Count}(k') / \text{Freq}(v)$.

\textbf{Step 3}. 
As mentioned previously, we hope that the original datastore $\mathcal{D}$ and the upstream NMT mode $\theta$ can also effectively model the above extracted semantically-related key-queries pairs via key representation revision, so as to make $\mathcal{D}$ more applicable to the downstream translation task. To this end, we traverse each extracted pair $(k', \{q'\})$ and determine their counterparts $(k, \{q\})$ using the datastore $\mathcal{D}$ and the model $\theta$. Note that $k$ and $k'$ are actually the hidden states at the same timestep, which are respectively generated by the models $\theta$ and $\theta '$ when traversing the same parallel sentence. Similarly, we determine the counterpart $q$ for $q'$. By doing so, we can obtain a set of key-queries pairs, denoted as $\mathcal{S}_r=\{(k,\{q\})\}$, as the training data of the reviser, where the key $k$ of each pair is expected to be semantically related to its corresponding queries in the semantic space of the original datastore.

\subsection{Training Objective}
With the above extracted key-queries pair set $\mathcal{S}_r$, we propose a training objective with two training losses to train the reviser:
\begin{align}
& \mathcal{L} = \sum_{(k,\{q\}) \in \mathcal{S}_r}(\mathcal{L}_{sd} + \alpha \mathcal{L}_{sc}),
\label{eq:overall-loss}
\end{align}
\noindent where $\alpha$ is a hyper-parameter that is used to control the effects of these two losses.

The first loss is the \textbf{semantic distance loss} $\mathcal{L}_{sd}$. Formally, given an extracted key-queries pair $(k, \{q\}) \in \mathcal{S}_r$, we define $\mathcal{L}_{sd}$ as follows:
\begin{align}
& \mathcal{L}_{sd} = d(k + \Delta k, \text{Avg}(\{q\})),
\label{eq:sd-loss}
\end{align}
\noindent where $\Delta k$ is the inductive bias vector produced by our reviser, and $\text{Avg}(\{q\})$ is the fixed average representation of extracted queries $\{q\}$. Note that $\mathcal{L}_{sd}$ constrains the direction of $\Delta k$. 
By minimizing this loss, the revised key representation is encouraged to approach the average representation of queries. In this way, the original datastore and upstream NMT model are also able to capture the key-queries semantic relevance revealed by the downstream datastore and NMT model.

However, it is widely known that a fine-tuned model often suffers from catastrophic forgetting~\cite{MCCLOSKEY1989109,ratcliff1990connectionist}. Likewise, if the key representations of the original datastore are significantly changed, they will forget a lot of translation knowledge learned by the upstream NMT model.

In order to avoid catastrophic forgetting, previous studies attempt to incorporate regularization relative to the original domain during fine-tuning~\cite{miceli-barone-etal-2017-regularization, kirkpatrick2017overcoming}. Inspired by these studies, we propose the second loss, called $\textbf{semantic consistency loss}$ $\mathcal{L}_{sc}$, to constrain the modulus of $\Delta k$: 
\begin{align}
& \mathcal{L}_{sc} = ||\Delta k||^2.
\label{eq:sc-loss}
\end{align}

Apparently, $\mathcal{L}_{sc}$ is essentially also a regularization term, which is used to retain the knowledge of the upstream NMT model by limiting the change of key representations.

\section{Experiments}
\begin{table*}[ht]
\resizebox{\linewidth}{!}{
\centering
\begin{tabular}{lcccccccccc}
\toprule
& \multicolumn{2}{c}{News} & \multicolumn{2}{c}{Koran} & \multicolumn{2}{c}{IT} & \multicolumn{2}{c}{Medical} & \multicolumn{2}{c}{Law} \\
\cmidrule(r){2-3} \cmidrule(r){4-5} \cmidrule(r){6-7} \cmidrule(r){8-9} \cmidrule(r){10-11}
& $k$NN-MT & Ours & $k$NN-MT & Ours & $k$NN-MT & Ours & $k$NN-MT & Ours & $k$NN-MT & Ours \\
\midrule
Koran & 20.31 & 21.28$\ddagger$ & - & - & 12.64 & 14.69$\ddagger$ & 9.51 & 10.79$\ddagger$ & 11.25 & 12.32$\ddagger$ \\
IT & 45.99 & 46.57$\dagger$ & 39.89 & 41.40$\ddagger$ & - & - & 29.06 & 30.82$\ddagger$ & 30.37 & 31.73$\ddagger$ \\
Medical & 54.12 & 55.77$\dagger$ & 50.66 & 52.55$\ddagger$ & 45.92 & 47.71$\ddagger$ & - & - & 46.96 & 49.14$\ddagger$ \\
Law & 61.27 & 61.77$\ddagger$ & 59.05 & 59.49$\ddagger$ & 44.82 & 46.22$\ddagger$ & 48.18 & 49.61$\ddagger$ & - & - \\
\midrule
Avg. & 45.42 & \textbf{46.35} & 49.87 & \textbf{51.15} & 34.46 & \textbf{36.21} & 28.92 & \textbf{30.41} & 29.53 & \textbf{31.06} \\
\bottomrule
\end{tabular}
}
\end{table*}

\begin{table*}[ht]
\resizebox{\linewidth}{!}{
\centering
\begin{tabular}{lcccccccccc}
\toprule
& \multicolumn{2}{c}{News} & \multicolumn{2}{c}{Koran} & \multicolumn{2}{c}{IT} & \multicolumn{2}{c}{Medical} & \multicolumn{2}{c}{Law} \\
\cmidrule(r){2-3} \cmidrule(r){4-5} \cmidrule(r){6-7} \cmidrule(r){8-9} \cmidrule(r){10-11}
& $k$NN-MT & Ours & $k$NN-MT & Ours & $k$NN-MT & Ours & $k$NN-MT & Ours & $k$NN-MT & Ours \\
\midrule
Koran & -0.183 & -0.163$\ddagger$ & - & - & -0.482 & -0.368$\ddagger$ & -0.717 & -0.639$\ddagger$ & -0.623 & -0.541$\dagger$ \\
IT & 0.524 & 0.526 & 0.394 & 0.455$\ddagger$ & - & - & -0.011 & 0.066$\ddagger$ & 0.054 & 0.100$\ddagger$ \\
Medical & 0.539 & 0.539 & 0.472 & 0.507$\ddagger$ & 0.304 & 0.348$\ddagger$ & - & - & 0.346 & 0.413$\ddagger$ \\
Law & 0.529 & 0.533$\dagger$ & 0.611 & 0.626$\ddagger$ & 0.184 & 0.232$\ddagger$ & 0.296 & 0.353$\ddagger$ & - & - \\
\midrule
Avg. & 0.352 & \textbf{0.359} & 0.492 & \textbf{0.529} & 0.002 & \textbf{0.071} & -0.144 & \textbf{-0.073} & -0.074 & \textbf{-0.009} \\
\bottomrule
\end{tabular}
}
% \vspace{-1mm}
\caption{
The ScareBLEU and COMET scores of the conventional $k$NN-MT model and ours on test sets, where all models are individually trained with an upstream training corpus and a downstream one, and then evaluated on multiple downstream test sets. The involved upstream and downstream domains are listed in the first row and the first column, respectively. \textbf{Bold} indicates the best result. $\dagger$ or $\ddagger$: significantly better than $k$NN-MT with t-test $p$<0.05 or $p$<0.01. Here we conduct 1,000 bootstrap tests~\cite{koehn-2004-statistical} to measure the significant difference between scores.
}
\label{tab:main-exp}
\end{table*}

To investigate the effectiveness of our method, we conduct experiments in the task of NMT domain adaptation.

\subsection{Settings}
\paragraph{Datasets and Evaluation}
\label{sec:dataset-setting}
We conduct experiments using the multi-domain datasets released by~\citet{aharoni-goldberg-2020-unsupervised}.
The details of these datasets are shown in Table~\ref{tab:dataset} of the Appendix. Unlike the previous studies~\cite{khandelwal2021nearest, zheng-etal-2021-adaptive, jiang2022towards} only using News as the upstream domain, we additionally use other available domains as upstream ones, which include Koran, IT, Medical, and Law.
We first use the Moses toolkit\footnote{https://github.com/moses-smt/mosesdecoder} to tokenize the sentences and split the tokens into subwords units~\cite{sennrich-etal-2016-neural}.
Finally, we use two metrics: case-sensitive detokenized BLEU~\cite{post-2018-call} and COMET~\cite{rei-etal-2020-comet}, to evaluate the quality of translation.

\paragraph{Baselines}
We select the following models as our baselines.
\begin{itemize}
% \vspace{-2mm}
\item \textbf{NMT}. When using News as the upstream domain, we directly use WMT'19 German-English news translation task winner~\cite{ng2019facebook} as the basic model. In the experiments with other upstream domains, we fine-tune this winner model on the corresponding upstream training corpus.
% \vspace{-2mm}
\item $k$\textbf{NN-MT}. It is a vanilla $k$NN-MT model, which is our most important baseline. It equips the conventional NMT model with a downstream datastore, where hyper-parameters are tuned on the corresponding development set.
\end{itemize}

\paragraph{Implementation Details}
Following~\citet{khandelwal2021nearest}, we adopt \emph{Faiss}~\cite{johnson2019billion} to conduct quantization and retrieval. As for the hyper-parameters of $k$NN-MT models including the weight $\lambda$ and temperature $T$, we directly use the setting of \cite{zheng-etal-2021-adaptive}. Besides, we set the number of retrieved pairs $N_k$ as 8 with Koran or IT as the downstream domain, and 4 otherwise. When filter pairs for the reviser training, we only retain 30\% extracted semantically-related key-queries pairs from the original datastore. The hidden size of the reviser is set as 8,192. When training this reviser, we empirically set the hyper-parameter $\alpha$ of the training objective (See Equation~\ref{eq:overall-loss}) to 2.0 in the experiments with upstream News domain, 0.4 for other experiments, and the number of training epoch as 100. During this process, we optimize the parameters using the Adam optimizer~\cite{kingma2014adam} with a learning rate of 5e-5.

\subsection{Effects of Hyper-parameter $\alpha$}
\begin{figure}[t]
\centering
\includegraphics[width=0.95\linewidth]{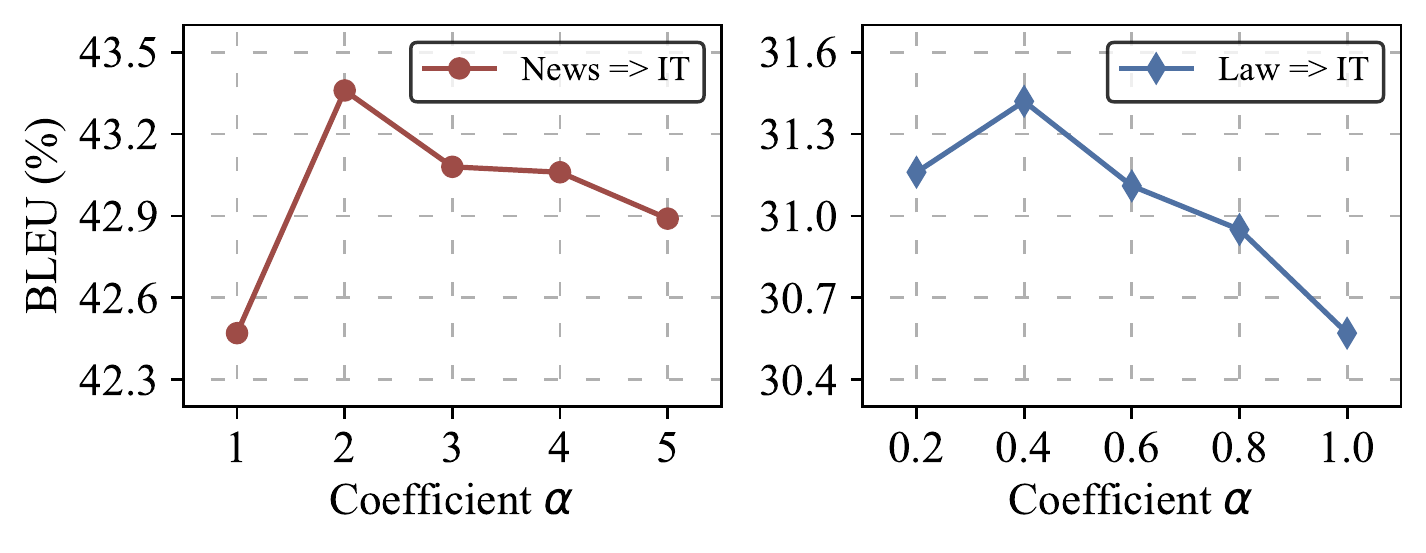}
\vspace{-3.5mm}
\caption{The ScareBLEU scores of our method with different coefficient $\alpha$ on News $\Rightarrow$ IT and Law $\Rightarrow$ IT (News $\Rightarrow$ IT indicates News is the upstream domain and IT is the downstream domain.).}
\label{fig:hyper-parameter-alpha}
\end{figure}

From Equation~\ref{eq:overall-loss}, we clearly know that the coefficient $\alpha$ is an important hyper-parameter controlling the effects of two losses. Hence, we first investigate its effects on our model.

Concretely, in the experiments with upstream News domain, we select IT as the downstream domain following previous studies~\cite{zheng-etal-2021-adaptive, wang-etal-2022-efficient}. Then, we explore the model performances with different $\alpha$ on the development set. The left subfigure of Figure~\ref{fig:hyper-parameter-alpha} illustrates the model performances with $\alpha$ varying from 1 to 5. We can find that our model achieves the best performance when $\alpha$ is 2.0. 
Therefore, we set $\alpha$ as 2.0 for all subsequent experiments using News as the upstream domain. In other groups of experiments, we still uniformly choose IT as the downstream domain, and Law as the upstream domain, where exists the largest amount of available data. We gradually vary $\alpha$ from 0.2 to 1.0 with an increment of 0.2, and also analyze the model performances on the corresponding development set. According to the experimental results reported in the right subfigure of Figure~\ref{fig:hyper-parameter-alpha}, we set $\alpha$ as 0.4 for all subsequent experiments with other upstream domains.

Notice that when setting News as the upstream domain, the optimal $\alpha$ is much larger than those of other upstream domains. 
As for this phenomenon, we speculate that the pre-trained NMT model of News domain involves large-scale training data and thus has learned more translation knowledge. Therefore, when applying our approach to experiments with upstream News domain, we set a relatively large $\alpha$ to effectively retain the translation knowledge of the pre-trained NMT model.

\subsection{Main Results}
Table~\ref{tab:main-exp} reports the performance of models on different domains. Overall, our model performs better than $k$NN-MT without introducing additional parameters in terms of two metrics. These results prove that our method is indeed able to effectively refine the $k$NN-MT datastore.

Specifically, in the experiments with upstream News domain, our model achieves only an average of +0.93 BLEU score on all domains, since the pre-trained NMT model for the upstream News domain is a competitive one and it involves the training data of other domains. Nevertheless, please note that this improvement is still significant at $p$<0.05. By contrast, in the experiments with other upstream domains, ours obtains more significant improvements.
\begin{table}[t]
\resizebox{\linewidth}{!}{
\centering
\begin{tabular}{lcccc}
\toprule
Upstream & \multicolumn{2}{c}{News} & \multicolumn{2}{c}{Law} \\
\cmidrule(r){2-3} \cmidrule(r){4-5}
Downstream & IT & Medical & IT & Medical \\
\midrule
Our method & \textbf{46.57} & \textbf{55.77} & \textbf{31.73} & \textbf{49.14} \\
\quad w/o data filtering & 45.24 & 54.70 & 31.56 & 48.82 \\
\quad w/o $\mathcal{L}_{sc}$ & 45.06 & 53.83 & 30.98 & 48.42 \\
\bottomrule
\end{tabular}
}
% \vspace{-2mm}
\caption{The ScareBLEU scores of ablation study.}
\label{tab:ablation}
\end{table}

\paragraph{Ablation Study}
To explore the effects of the data filtering strategy (See Section~\ref{sec:construction}) and $\mathcal{L}_{sc}$ (See Equation~\ref{eq:sc-loss}) on our model, we provide the performance of two variants of our model: 1) w/o data filtering. During the process of training the reviser, we do not filter any key-queries pairs extracted from the downstream datastore by the downstream NMT model. 2) w/o $\mathcal{L}_{sc}$. We only use the semantic distance loss to train the reviser for this variant. Following previous studies~\cite{zheng-etal-2021-adaptive, wang-etal-2022-efficient}, we consider News and Law as upstream domains and select IT and Medical as downstream domains.
In Figure~\ref{fig:domain-difference} of the Appendix, we find that these two domains are least related to News and Law. As shown in Table~\ref{tab:ablation}, the removal of the data filtering strategy or $\mathcal{L}_{sc}$ leads to a performance decline, proving the effectiveness of our model.

\subsection{Analysis}
\paragraph{Performance Improvement vs. Domain Difference}
\begin{figure}[t]
\centering
\includegraphics[width=0.95\linewidth]{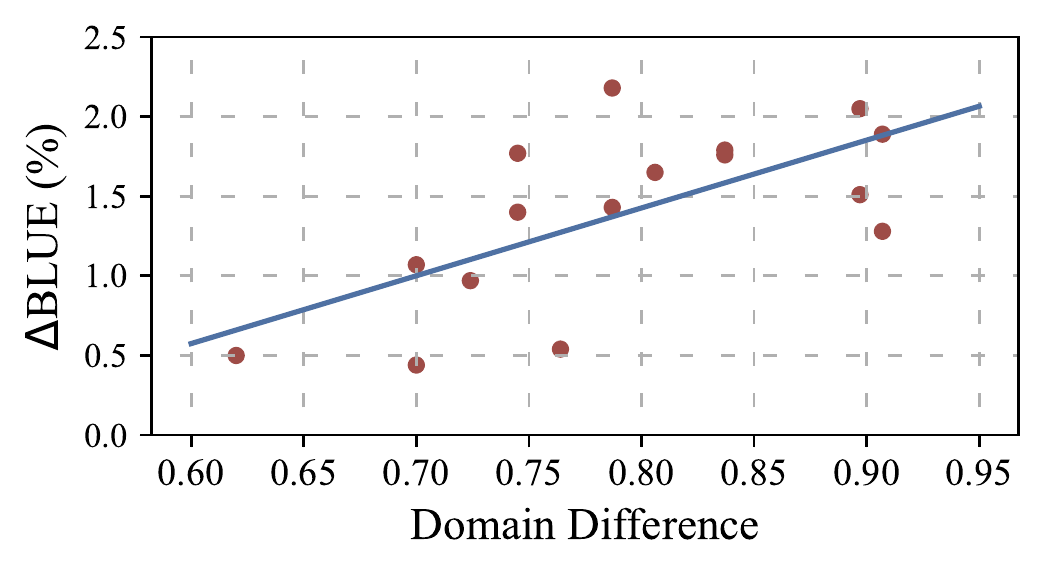}
\vspace{-2.5mm}
\caption{The domain differences for domain pairs and their corresponding performance improvements ($\Delta$BLEU).}
\label{fig:domain-difference}
\end{figure}
To further verify the rationality of our method, we explore the correlation between the performance improvements brought by our method and domain differences.
To this end, following~\citet{aharoni-goldberg-2020-unsupervised}, we first represent each domain with the average TF-IDF representation of its sentences on the development set, and then measure the domain difference according to the cosine similarity based on domain representations: $\text{Diff}(d_1, d_2) = 1 - \text{Cosine}(d_1, d_2)$.
In Figure~\ref{fig:domain-difference}, we plot the domain difference value and performance improvement for each domain pair. Here, we can observe a general trend that the greater the domain difference is, the more significant the performance improvement can be achieved by our method. Moreover, we measure Pearson's correlation coefficient between domain differences and performance improvements, resulting in a strong correlation value of 0.66\footnote{Given the significance level of 0.01 and the sample size of 16, the corresponding critical Pearson’s correlation value is 0.59.}.
These results prove the rationality of our method, and may also guide the evaluation of performance improvements of our approach in unseen domain pairs.

\paragraph{Compatibility of Our Method with Adaptive $k$NN-MT}
\begin{table}[t]
\resizebox{\linewidth}{!}{
\centering
\begin{tabular}{lcccc}
\toprule
Upstream & \multicolumn{2}{c}{News} & \multicolumn{2}{c}{Law} \\
\cmidrule(r){2-3} \cmidrule(r){4-5}
Downstream & IT & Medical & IT & Medical \\
\midrule
$k$NN-MT & 45.99 & 54.12 & 30.37 & 46.96 \\
Ours & 46.57 & 55.77 & 31.73 & 49.14 \\
\midrule
Adaptive $k$NN-MT & 47.51 & 55.87 & 31.52 & 48.43 \\
\quad + Ours & \textbf{47.99} & \textbf{56.27} & \textbf{32.64} & \textbf{49.67} \\
\midrule
Robust $k$NN-MT & 48.69 & 56.89 & 32.12 & 49.97 \\
\quad + Ours & \textbf{49.12} & \textbf{57.25} & \textbf{34.05} & \textbf{50.81} \\
\bottomrule
\end{tabular}
}
% \vspace{-2mm}
\caption{
The ScareBLEU scores of Adaptive $k$NN-MT~\cite{zheng-etal-2021-adaptive} and Robust $k$NN-MT~\cite{jiang2022towards} with the datastore revised by our method.}
\label{tab:compatible-exp}
\end{table}
As one of the most commonly-used $k$NN-MT variants, Adaptive $k$NN-MT~\cite{zheng-etal-2021-adaptive} dynamically estimates the weight $\lambda$ for $k$NN-MT to filter noises. Along this line, Robust $k$NN-MT~\cite{jiang2022towards} incorporates the confidence of NMT prediction into the dynamic estimation of $\lambda$, achieving further improvement.
Noteworthy, Adaptive $k$NN-MT, Robust $k$NN-MT, and our approach are able to alleviate the negative effects of the domain gap on $k$NN-MT from different perspectives.
Furthermore, we explore whether our method is compatible with Adaptive $k$NN-MT and Robust $k$NN-MT. To ensure a fair comparison, we use the same retrieval number for Adaptive $k$NN-MT.
From Table~\ref{tab:compatible-exp}, we can observe that the performance of Adaptive $k$NN-MT and Robust $k$NN-MT can be further improved with our approach.

\paragraph{Retrieval Accuracy}
\begin{table}[t]
\fontsize{9}{11}\selectfont
\setlength{\tabcolsep}{1.1mm}
\centering
\begin{tabular}{lcccc}
\toprule
Upstream & \multicolumn{4}{c}{News}\\
\cmidrule(r){2-5}
Downstream & Koran & IT & Medical & Law \\
\midrule
$k$NN-MT & 41.43 & 62.45 & 72.56 & 79.85 \\
Ours & \textbf{44.87} & \textbf{63.92} & \textbf{74.18} & \textbf{81.45} \\
\bottomrule
\end{tabular}
% \vspace{-2mm}
\caption{
The retrieval accuracy of $k$NN-MT and our model on the experiment with upstream News domain.}
\label{tab:retrieval-accuracy}
\end{table}
To verify the effectiveness of our method on datastore retrieval, we analyze the retrieval accuracy of the $k$NN-MT model with or without our strategy. As shown in~\ref{tab:retrieval-accuracy}, our method always achieves higher retrieval accuracy than the conventional $k$NN-MT. It indicates that the performance improvement of our method comes from the improvement of datastore quality.

\paragraph{Effects of Hyper-parameter $r$}
To demonstrate the effectiveness of our method, we also explore the effect of the hyper-parameter: the selected percentage $r$\% of collected semantically-related key-queries pairs when constructing training data. As shown in Table~\ref{tab:hyper-parameter-r}, we find that our method outperforms $k$NN-MT with various $r$\%. Besides, with the percentage $r$\% increasing, the performance of our method can be further improved. In practice, we set $r$\% as 30\% to balance the training resource overhead and performance improvement.
\begin{table}[t]
\fontsize{9}{11}\selectfont
\setlength{\tabcolsep}{1.1mm}
\centering
\begin{tabular}{lcccc}
\toprule
Upstream & \multicolumn{4}{c}{News}\\
\cmidrule(r){2-5}
Downstream & Koran & IT & Medical & Law \\
\midrule
$k$NN-MT & 20.31 & 45.99 & 54.12 & 61.27 \\
Ours ($r$ = 20) & 21.12 & 46.34 & 55.42 & 61.48 \\
Ours ($r$ = 30) & 21.28 & 46.57 & \textbf{55.77} & 61.77 \\
Ours ($r$ = 40) & \textbf{21.30} & \textbf{46.90} & 55.51 & \textbf{61.82} \\
\bottomrule
\end{tabular}
\caption{
The ScareBLEU scores of our method with different percentages $r$\% of data retention on test sets.}
\label{tab:hyper-parameter-r}
\end{table}

\subsection{Discussion}
\paragraph{Our Method vs. Fine-tuning}
As mentioned in Section~\ref{sec:construction}, our method use the downstream NMT model to construct training data, where the downstream NMT model is obtained by fine-tuning the upstream NMT model on the downstream training corpus.
Despite the requirement for more training resources, our method has a significant advantage in deploying resource overhead (see Section~\ref{sec:introduction}). Besides, our method still retains the following advantages of conventional $k$NN-MT: 1) Interpretable. This is because the retrieval process of $k$NN-MT is inspectable, the retrieved highly-relevant examples can be directly traced back to the specific sentence in the training corpus; 2) Flexible. We can use arbitrary amounts of data to build the datastore, and thus we can increase or decrease the amount of data in the datastore at will as needed immediately.

\section{Related Work}
Our related work mainly includes two aspects: domain adaptation for NMT, and non-parametric retrieval-augmented approaches for NMT.

\paragraph{Domain Adaptation for NMT}
As summarized in \citet{chu-wang-2018-survey}, dominant methods in this aspect can be roughly divided into two categories: 1) model-centric approaches that focus on carefully designing NMT model architecture to learn target-domain translation knowledge~\cite{wang-etal-2017-instance, zeng-etal-2018-multi, bapna2019non, guo-etal-2021-parameter}, or refining the training procedures to better exploit context~\cite{wuebker2018compact, bapna-firat-2019-simple, lin-etal-2021-learning, liang2021finding}; 2) data-centric methods resorting to leveraging the target-domain monolingual corpus~\cite{zhang-zong-2016-exploiting, zhang2018joint}, synthetic corpus~\cite{hoang2018iterative, hu-etal-2019-domain-adaptation, wei-etal-2020-iterative} or parallel corpus~\cite{chu-etal-2017-empirical} to improve the NMT model via fine-tuning. 

\paragraph{Non-parametric Retrieval-augmented Approaches for NMT}
Generally, these methods retrieve sentence-level examples to enhance the robustness and expressiveness of NMT models~\cite{zhang-etal-2018-guiding, bulte-tezcan-2019-neural, xu-etal-2020-boosting}. For example, \citet{zhang-etal-2018-guiding} retrieves similar source sentences with target tokens from a translation memory, which are used to increase the probabilities of the collected tokens. Both \citet{bulte-tezcan-2019-neural} and \citet{xu-etal-2020-boosting} use the parallel sentence pairs retrieved via fuzzy matching as the auxiliary information of the current source sentence. 

\cite{khandelwal2021nearest} is the first attempt to explore \emph{k}NN-MT, showing its effectiveness on non-parametric domain adaptation for NMT. Following this work, researchers have proposed \emph{k}NN-MT variants, which mainly include two research lines: 1) the first line is mainly concerned with accelerating model inference by adaptive retrieval~\cite{he-etal-2021-efficient}, datastore compression~\cite{he-etal-2021-efficient, wang-etal-2022-efficient, martins-etal-2022-efficient}, or limiting the search space by source tokens~\cite{meng-etal-2022-fast}; 2) the second line focuses on reducing noises in retrieval results, through dynamically estimating the hyper-parameter $N_k$ or the interpolation weight $\lambda$~\cite{jiang-etal-2021-learning, zheng-etal-2021-adaptive, wang2022non, jiang2022towards}.
In addition, \citet{zheng-etal-2021-non-parametric} present a framework that uses downstream-domain monolingual target sentences to construct datastores for unsupervised domain adaptation.

Unlike the above studies caring more about filtering noise in retrieval results, inspired by representation learning~\cite{su-etal-2015-bilingual, su-etal-2016-convolution, Zhang_Xiong_Su_2017}, we are mainly concerned with enhancing $k$NN-MT by revising the key presentations of the datastore.
Note that very recently, \citet{wang-etal-2022-learning-decoupled} use an adapter to generate better retrieval representations in an online manner. 
However, unlike this work, we revise the key representation of the $k$NN-MT datastore in an offline manner. 
Besides, our method does not introduce additional parameters during inference, and thus maintains resource overhead.

\section{Conclusion}
In this paper, we first conduct a preliminary study to investigate the impact of the domain gap on the datastore retrieval of $k$NN-MT.
Furthermore, we propose a reviser to refine the key representations of the original $k$NN-MT datastore in an offline manner, 
making them more suitable for the downstream domain. 
This reviser is trained on the collection of key-queries pairs, 
where the key of each pair is expected to be semantically related to its corresponding queries. 
Particularly, we introduce two losses to train the reviser, ensuring that the revised key representations conform to the downstream domain while effectively retaining their original knowledge. 
Through extensive experiments, we demonstrate the effectiveness of our method. 
Besides, in-depth analyses reveal that: 1) the performance improvement achieved by our method is positively correlated with the degree of the domain gap; 2) this improvement is primarily attributed to the enhancement of the datastore quality; 3) our method is able to compatible with existing Adaptive $k$NN-MT.

To further verify the generalization of our method, we will extend our method to $k$NN-LM or other text generation tasks, such as controllable generation.

\section*{Limitations}
When using our method, we have to fine-tune the upstream NMT model to construct the downstream NMT model and then datastore for the reviser training. Hence, compared with the current commonly-used $k$NN-MT variant~\cite{zheng-etal-2021-adaptive}, our method requires more time for training. Nevertheless, it does not introduce additional parameters during inference.

\section*{Acknowledgements}
The project was supported by National Natural Science Foundation of China (No. 62036004, No. 62276219), Natural Science Foundation of Fujian Province of China (No. 2020J06001), Youth Innovation Fund of Xiamen (No. 3502Z20206059), and Alibaba Group through Alibaba Innovative Research Program. We also thank the reviewers for their insightful comments.

\bibliography{anthology,custom}

\appendix
\newpage
\section{Dataset Statistics}
\begin{table}[!h]
\fontsize{10}{12}\selectfont
\setlength{\tabcolsep}{1.1mm}
\centering
\begin{tabular}{lccccc}
\toprule
 & Koran & IT & Medical & Law \\
\midrule
Train & 18K & 223K & 248K & 467K \\
Dev & 2K & 2K & 2K & 2K \\
Test & 2K & 2K & 2K & 2K \\
\bottomrule
\end{tabular}
\caption{
The example numbers of training, development, and test sets in four domains.
}
\label{tab:dataset}
\end{table}

\section{Domain Difference}
\begin{figure}[!h]
\centering
\includegraphics[width=0.98\linewidth]{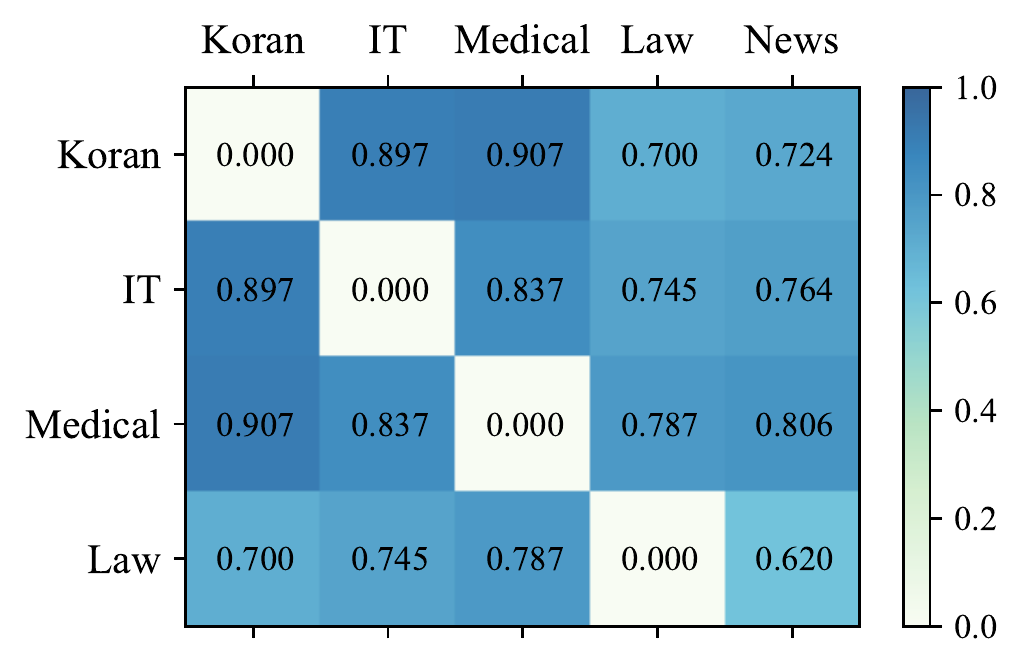}
\caption{Domain Difference for each domain pair. The darker color denotes the greater difference.}
\label{fig:domain-difference}
\vspace{-4mm}
\end{figure}

\section{The Effect of Hyper-Parameter $N_k$}
\begin{table}[!h]
\fontsize{9}{11}\selectfont
\setlength{\tabcolsep}{1.1mm}
\centering
\begin{tabular}{lcccc}
\toprule
News $\Rightarrow$ IT & $N_k$ = 4 & $N_k$ = 8 & $N_k$ = 12 & $N_k$ = 16 \\
\midrule
$k$NN-MT & 44.77 & 45.99 & 45.34 & 45.25 \\
Ours & \textbf{45.40} & \textbf{46.57} & \textbf{45.88} & \textbf{45.63} \\
\bottomrule
\end{tabular}
\caption{
The ScareBLEU scores of our method with different retrieve pairs $N_k$ on News $\Rightarrow$ IT.}
\label{tab:hyper-parameter-nk}
\end{table}
To demonstrate the reliability of our method, we also explore our method with different hyper-parameter $N_k$. As shown in Table~\ref{tab:hyper-parameter-nk}, our method enjoys consistent performance under different $N_k$.

\end{document}